\documentclass[runningheads]{llncs}
\usepackage{amssymb}
\usepackage[T1]{fontenc}
\usepackage{dsfont}
\usepackage{amsmath} 
\usepackage{array}
\usepackage{multirow}
\usepackage{colortbl}

\newcolumntype{C}[1]{>{\centering\arraybackslash}p{#1}}
\usepackage{graphicx,verbatim}
\usepackage{hyperref}

\begin{document}
\title{NoduleFlow: Automatic 3D Lung Nodule Detection and Segmentation}


\titlerunning{NoduleFlow}
%

\author{\large Giulia Romoli$^{1,*}$, Filippo Ruffini$^{1,2}$, Francesco Di Feola$^{1}$, Valerio Guarrasi$^{2}$, Matteo Tortora$^{3}$ and Paolo Soda$^{1,2,*}$}
\authorrunning{Giulia Romoli et al.}
\institute{\footnotesize $\ $\\$^1$ Department of Diagnostics and Intervention, Radiation Physics, Biomedical Engineering, Umeå University, Universitetstorget
4, Umeå, Sweden. \\
$^2$ Unit of Artificial Intelligence and Computer Systems, Department of Engineering, Università Campus Bio-Medico di Roma, Via Alvaro del
Portillo 21, Rome, Italy. \\
$^3$ Department of Naval, Electrical, Electronic, and Telecommunications Engineering, University of Genoa, Via Opera Pia 11A, Genoa, Italy.  \\
$^*$corresponding author: giulia.romoli@umu.se, paolo.soda@umu.se
}
  
\maketitle              
\begin{abstract}
Low-dose Computed Tomography (CT) is the standard imaging modality for lung cancer screening. 
On CT scans, lung cancer typically manifests as pulmonary nodules, whose manual identification is time-consuming and prone to observer variability, image quality limitations, and suboptimal reading conditions. 
Moreover, subtle temporal changes are often difficult to quantify reliably in CT scans from the same patient across follow-up examinations. 
Deep learning–based approaches for the automatic detection and segmentation of lung nodules have demonstrated improved accuracy and efficiency. 
However, most existing approaches provide only coarse region-of-interest localization, rely on prior lesion annotations or human supervision, or treat detection and segmentation as separate tasks. 
In this work, we present NoduleFlow, a framework for lung nodule detection, classification, and segmentation directly from series of CT volumes. Our proposal consists of four interconnected modules and operates in an automated manner at inference time, without requiring prior lesion annotations, manual initialization, or region-of-interest pre-selection. We further investigate whether nodule localization benefits from the propagation of candidate regions across follow-up scans through inter-scan registration fields, as opposed to conventional independent per-scan analysis.
The generalizability of the framework is systematically evaluated across four independent datasets. The results underscore the potential of automated pipelines for lung nodule analysis in screening and diagnostic settings.


\end{abstract}

\section{Introduction}

Lung cancer remains the leading cause of cancer-related mortality worldwide~\cite{bray2024}, for which low-dose Computed Tomography (CT) is the standard imaging modality for screening~\cite{national2011,kovalchik2013,difeola2025}. 
On CT scans, primary lung cancer typically manifests as a pulmonary nodule, whose manual identification is time-consuming and subject to observer variability. 
Missed detections may arise from nodule characteristics, image quality limitations, or fatigue, distraction, and suboptimal reading conditions~\cite{zhong2024,melzer2024}. 
Furthermore, comparison of nodules across serial CT examinations can reveal nodule growth patterns, yet subtle temporal changes are often difficult to quantify robustly.
These limitations have driven interest in tools based on Artificial Intelligence (AI) for computer-aided diagnosis (CAD) to reduce radiologists' workload and advance lung cancer diagnosis. 
A typical pipeline for lung cancer diagnosis is composed of three main stages: pulmonary nodule detection, classification, and segmentation~\cite{gayap2024,quanyang2024}. 
Detection algorithms identify candidate regions of interest (ROIs) for potential lung nodules, and generally incorporate False Positive (FP) reduction strategies to manage the trade-off between high sensitivity and acceptable specificity.
Classification models distinguish between benign and malignant nodules. 
Finally, segmentation algorithms compute a mask that delineates the nodule and separates it from surrounding tissues. When multiple CTs from the same patient acquired during follow-up are available, registration models estimate deformation fields that spatially align individual scans.

Despite progress, existing solutions still exhibit limitations. ROIs from detection models provide only coarse nodule localization and lack precise delineation of lesion boundaries. 
Nodule classification and segmentation methods often rely on human supervision or prior lesion localization. 
Registration ensures spatial consistency across multiple CT scans, at the expense of an increased computational load and the introduction of potential interpolation artifacts.
Several approaches work on 2D images neglecting the intrinsic volumetric information~\cite{marinakis2024}. Moreover, most existing AI-based solutions analyze each CT scan independently, discarding the longitudinal imaging history that is critical for more reliable malignancy risk stratification, the monitoring of treatment response, and the detection of newly emerging lesions.
Only a limited number of studies have proposed 3D end-to-end pipelines that jointly address detection and segmentation without requiring external inputs~\cite{tang2019,ardimento2023,kashyap2025}. 
To the best of our knowledge, none of the end-to-end solutions proposed so far incorporates longitudinal modeling, i.e., the joint analysis of serial CT exams from the same patient to capture temporal dependencies and patterns between scans.
They process each CT scan independently, without exploiting complementary information of the patient from follow-up examinations. 
Methods that leverage longitudinal CT scans remain dependent on strong prior assumptions or partial supervision. Among the approaches proposed so far, some require external prompts at the baseline time~\cite{hering2021,rokuss2025}, others assume that lesions are already segmented across time~\cite{diveroli2023,fernandes2024}, or are limited to track point locations rather than performing full volumetric segmentation~\cite{tang2022,vizitiu2023,ke2022}. 

This work presents an AI framework for the automatic detection, classification, and segmentation of lung nodules from series of CT volumes. 
The main contributions are summarized as follows: 
i) we propose an automated AI framework for lung nodule identification without reliance on prior external prompts at inference time; 
ii) we investigate whether exploiting a longitudinal propagation strategy of ROIs across follow-up scans can improve nodule localization compared to independent per-scan analysis; 
iii) the generalizability of the proposed framework is systematically evaluated across four independent datasets, demonstrating robustness to dataset-specific variations.

\section{Methods and Materials}

\begin{figure}
\includegraphics[width=\textwidth]{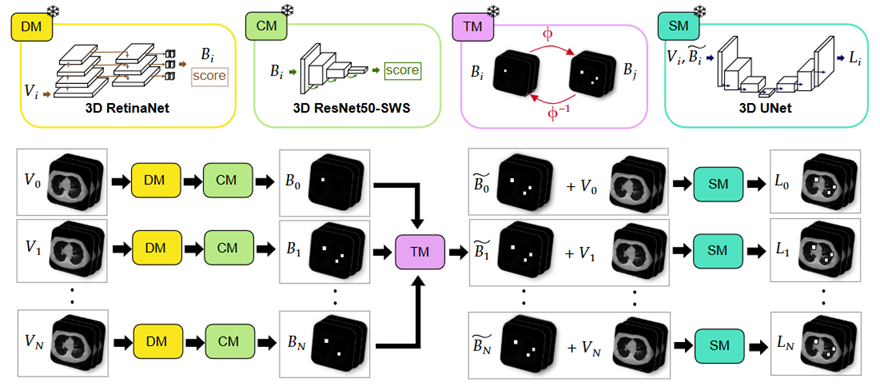}
\caption{Overview of the proposed framework. For each CT scan $V_i$, a Detection Module (DM) and a Classification Module (CM) generate a binary volumetric ROI $B_i$ for candidate nodule locations. Volumetric ROIs are then processed by a Transfer Module (TM) to produce a refined ROI $\widetilde{B_i}$, which, together with the original CT scan $V_i$, is passed to the Segmentation Module (SM) to obtain the nodule segmentation mask $L_i$. } \label{fig1}
\end{figure}

Our framework integrates detection, classification, and segmentation of lung nodules from series of CT volumes. 
To address each stage of this workflow, the framework is organized into four pre-trained modules (Fig.~\ref{fig1}): a Detection Module (DM) that localizes candidate nodules, a Classification Module (CM) that filters them based on a malignancy probability, a Transfer Module (TM) that propagates candidate locations across follow-up time points, and a Segmentation Module (SM) that computes precise volumetric segmentations.

The Detection Module (DM) is a 3D RetinaNet~\cite{lin2017} that processes each CT volume $V_i$ at time $i$ independently, producing a set of $K$ candidate detections $B_i = \{(b_{ik}, s_{ik})\}_{k=1}^{K}$, where $b_{ik} \in \mathbb{R}^6$ is the bounding box and $s_{ik} \in [0,1]$ its detection confidence score. 

The Classification Module (CM) is a 3D ResNet50-SWS~\cite{tushar2024} that associates to each candidate an additional malignancy probability score $m_{ik} \in [0,1]$, yielding an updated set $B_i = \{(b_{ik}, s_{ik}, m_{ik})\}_{k=1}^{K}$. 
As a false-positive reduction strategy, spatially overlapping candidates are merged into a single detection. Both confidence and malignancy scores are recomputed from the $J$ overlapping candidates via $s_i = 1 - \prod_{j=1}^{J}(1 - s_{ij})$ and $m_i = 1 - \prod_{j=1}^{J}(1 - m_{ij})$ respectively.
The final score increases only when multiple overlapping candidates yield high score values, whereas isolated or low-score detections are penalized.

The Transfer Module (TM) aggregates detection bounding boxes across time using GradICON~\cite{tian2023,tian2024}, a multi-resolution cascade of U-Nets that estimates diffeomorphic transformations $\Phi$ between image pairs, satisfying $\Phi\Phi^{-1}\approx \mathds{1}$. 
Diffeomorphic registration preserves topological structure and guarantees bidirectional consistency, making it particularly suited for robust longitudinal modeling.
Within TM, GradICON is applied to each pair $(V_i, V_j)$, $i \neq j$, estimating a displacement field $\Phi_{ji}: \mathbb{R}^3 \rightarrow \mathbb{R}^3$ that maps coordinates from $V_i$ to $V_j$. 
The resulting field $\Phi_{ji}$ is used to propagate each bounding box $b_{ik}$ from $V_i$ to $V_j$, yielding the warped candidate $\tilde{b}_{jk} = \Phi_{ji}\cdot{b_{ik}}$ in $V_j$. 

We introduce three alternative strategies to aggregate propagated candidates across time points, producing a refined detection set $\widetilde{B}_i = \{(\tilde{b}_{ik}, s_{ik}, m_{ik})\}_{k=1}^{\tilde{K}}$ for each time point $i$, where $\tilde{K}_i$ denotes the number of candidates after merging.
A baseline strategy (S0) considers no temporal propagation, retaining independent detections at each time point, i.e., $\widetilde{B}_i = B_i$. Strategy (S1) aggregates candidates from prior examinations only, leveraging longitudinal context while avoiding the use of future information:
\begin{equation}
\widetilde{B}_i = \bigcup_{t \leq i} \Phi_{it}\cdot{B_t} = \{(\Phi_{it}\cdot{b_{tk}}, s_{ik}, m_{ik})\}_{k=1}^{\tilde{K}} 
\end{equation}

\noindent
where $\Phi_{it}$ denotes the displacement field mapping coordinates from $V_t$ to $V_i$. 

Strategy (S2) aggregates candidates from all available time points:
\begin{equation}
\widetilde{B}_i = \bigcup_{t} \Phi_{it}\cdot{B_t} = \{(\Phi_{it}\cdot{b_{tk}}, s_{ik}, m_{ik})\}_{k=1}^{\tilde{K}} 
\end{equation}

While less suitable for clinical deployment, S2 is particularly advantageous for retrospective analyses where the complete longitudinal series is accessible.

The last part of the framework is a Segmentation Module (SM). 
SM is a 3D UNet~\cite{isensee2024} and returns segmentations of lung nodules $L_i$ directly from the original CT volumes $V_i$, using the refined bounding boxes provided by the TM module as annotations. 
SM supports both bounding box and point annotations, the last computed as the geometric centers of the corresponding bounding boxes.

The pipeline integrates publicly available, independently pre-trained models, assembled sequentially without any joint training, fine-tuning, or task-specific adaptation. 
While individual modules address well-established tasks, the proposed framework contributes their orchestration into a fully automated pipeline and provides a controlled setting to evaluate whether longitudinal information can enhance nodule localization. 
Specifically, the TM bridges detection and segmentation by propagating candidate ROIs across time points through registration fields, enabling the segmentation module to potentially benefit from longitudinal context without operating on spatially aligned volumes. 
This design preserves nodule morphology, avoids deformation-induced biases, and enables accurate volumetric and mass estimation while still benefiting from longitudinal contextual information encoded at the detection level.

For the DM module, we employed two variants of the 3D RetinaNet model, separately trained on the Duke Lung Cancer Screening (DLCS)~\cite{wang2025} and LUNA16 ~\cite{armato2011} datasets. 
We additionally evaluated an ensemble detector that combines the outputs of both models, rewarding mutually consistent predictions following the same approach used for confidence and malignancy scores. 
For the CM module, we selected the 3D ResNet50-SWS architecture, which demonstrated superior performance when compared against a benchmark of alternative classification models~\cite{tushar2024}.
For the SM module, we employed 
LesionLocator~\cite{rokuss2025}, a promptable 3D UNet segmentation architecture. 
All components were used in their original, frozen configuration as released by their respective authors. 

We used four independent datasets for a total of 481 patients, 1266 scans, and 4593 annotated lung nodules as test sets to evaluate the framework. 
From the National Lung Screening Trial (NLST), we consider two annotated subsets of patients having one or more follow-up scans: NLST-Sybil~\cite{mikhael2023}, from which we selected a cohort of 146 patients with 398 nodules, and NLST-LongCT~\cite{LongCT}, including 97 patients with 110 annotated nodules.
From the Longitudinal-CT dataset~\cite{LongitudinalCT}, which contains paired baseline and post-treatment scans of melanoma patients, we retained 169 patients with 2166 annotated lung metastases. 
Finally, from the UniToChest dataset~\cite{chaudhry2022}, we selected a longitudinal subset of 69 patients with 1919 annotated nodules.
We preprocessed all datasets using a standardized pipeline.  
CT scans were cropped to the lung region using masks generated by TotalSegmentator~\cite{wasserthal2023}. 
All follow-up CT volumes $V_{i>0}$ were processed to match the spatial resolution and image dimensions of the corresponding baseline $V_0$. 

We derived lesion diameters D from ground-truth annotations and used them to analyze performance metrics as a function of lesion size, where D is defined as the average of its maximal length and maximal orthogonal diameter on axial CT slice~\cite{macmahon2017}. 
For the computation of NLST diameters, we generated lesion masks using LesionLocator~\cite{rokuss2025} from the available ground-truth annotations.

\section{Results and Discussion}

We evaluate detection using the Competition Performance Metric (CPM)~\cite{setio2017}, defined as average sensitivity at 1/8, 1/4, 1/2, 1, 2, 4, and 8 FP per scan. 
For nodule-level cancer classification, we use the receiver operating characteristic area under the curve (AUC). 
Finally, we assess segmentation using metrics appropriate to the available ground-truth annotations in each dataset. 
For Longitudinal-CT and UniToChest, where voxel-level reference masks are available, we report the Dice Similarity Coefficient (DSC). 
For NLST-Sybil, where annotations consist of bounding boxes, we report Intersection over Union (IoU) between predicted and ground-truth bounding boxes. 
For NLST-LongCT, where only point annotations are available, we report the hit rate, defined as the proportion of annotated points falling within a predicted segmentation mask. 

Results are compared against TotalSegmentator~\cite{wasserthal2023} (TS), an nnU-Net framework that automatically segments lung nodules on CT scans; 
NoduleNet (NNet) ~\cite{tang2019}, a 3D deep Convolutional Neural Network for nodule detection and segmentation;
and MedSAM2~\cite{ma2025medsam2} (MS2), a prompt-driven segmentation model derived from the Segment Anything Model paradigm. 
Among the 3D end-to-end approaches proposed in the literature~\cite{tang2019,ardimento2023,kashyap2025}, NNet is the only one whose implementation was publicly released along with training data. We trained NNet on the LIDC-IDRI dataset~\cite{LIDC-IDRI} and compared it with our approach. 

\textbf{DM performance:} Table~\ref{tab1} reports the CPM and sensitivity at 8 FP/scan for the three variants of the DM, evaluated across the test datasets and compared with NoduleNet (NNet): the two 3D RetinaNet models separately trained on DLCS and LUNA16, and their ensemble (D-De, L-De, and ENS).
Across most datasets, the ensemble detector achieved the highest CPM, reaching 70.9\% on NLST-Sybil and 36.8\% on NLST-LongCT, outperforming NNet (44.7\%
and 27.0\%, respectively). However, NNet remained superior on LongCT (38.0\% vs. 19.0\%) and UniToChest (28.1\% vs. 25.9\%).  
We observe that the detection performances are strongly influenced by lesion size and the presence of metastases, with numerous small and spatially clustered nodules that increase the likelihood of missed detections. 
To deepen this issue, we provide results on subsets of Longitudinal-CT and UniToChest restricted to scans with less than 6 lesions per volume (Ls/scan), motivated by the clinical distinction between oligometastatic and polymetastatic disease. In UniToChest, when restricting the detection to scans with at most 6 Ls/scan, performance increases from $25.9\%$ to $72.0\%$. Indeed, the median diameter of this dataset increases from $5\;\text{mm}$ (IQR: $3$–$8\;\text{mm}$) to $12\;\text{mm}$ (IQR: $6$–$22\;\text{mm}$), suggesting that larger and easier-to-detect lesions predominate once highly metastatic cases are excluded. In contrast, Longitudinal-CT performance remains suboptimal even after restriction ($19.0\%$ and $26.9\%$), due to persistently small lesions which are inherently more challenging for detection.
While we report results for all three detector variants, subsequent modules (CM, TM, SM) were evaluated using the ensemble output.

\textbf{CM performance:} Nodule malignancy classification results from the 3D ResNet50-SWS model gained an AUC of $92.1\%$, $95.6\%$, $90.4\%$, and $90.4\%$ for NLST-Sybil, NLST-LongCT, LongitudinalCT, and UniToChest respectively.

\begin{table}[!h]
\centering
\caption{Average CPM ($\%$) and sensitivity at 8 FP/scan (SEN, $\%$) of the detection models (D-De, L-De, and ENS) on the test datasets, compared to NoduleNet (NNet). For Longitudinal-CT and UniToChest, results are also given for a subset of scans with less than 6 lesions (Ls/scan).  Bold indicates best performance. }\label{tab1}
\begin{tabular}{|c|c|c|c|c|c||c|c|c|c|}
\cline{3-10}
\multicolumn{2}{c|}{} &
\multicolumn{4}{c||}{\textbf{CPM} \(\uparrow\)} &
\multicolumn{4}{c|}{\textbf{SEN (8FP/scan) \(\uparrow\) }} \\
\hline

\fontsize{8}{0} \selectfont \textbf{Dataset} & \fontsize{8}{0} \selectfont \textbf{Selection} & \fontsize{8}{0} \selectfont D-De & \fontsize{8}{0} \selectfont L-De & \fontsize{8}{0} \selectfont ENS & \fontsize{8}{0} \selectfont NNet\cite{tang2019}&\fontsize{8}{0} \selectfont D-De & \fontsize{8}{0} \selectfont L-De & \fontsize{8}{0} \selectfont ENS & \fontsize{8}{0} \selectfont NNet\cite{tang2019} \\
\hline

\multirow{2}{*}{\fontsize{8}{0} \selectfont \textbf{LongCT}} & all    & 14.6 & 15.8 & 19.0 & \textbf{38.0} & 21.2 & 24.8 & 26.6 & \textbf{66.9}\\
& \fontsize{8}{0} \selectfont ${\leq}6$ Ls/scan & 23.1 & 21.8 & 26.9 & \textbf{46.1}& 30.3 & 33.9 & 36.4 & \textbf{71.3} \\
\hline

\multirow{2}{*}{\fontsize{8}{0} \selectfont \textbf{UniToChest}} & all  & 18.1 & 22.5 & 25.9 & \textbf{28.1} & 22.5 & 28.0 & 30.3 & \textbf{47.6}\\
& \fontsize{8}{0} \selectfont ${\leq}6$ Ls/scan & 66.3 & 60.6 & \textbf{72.0} & 47.4 & 69.8 & 68.2 & \textbf{75.7} & 65.4 \\
\hline
\fontsize{8}{0} \selectfont \textbf{NLST-Sybil}   & all & 67.6 & 48.0 & \textbf{70.9} & 44.7 & 76.1 & 63.6 & \textbf{78.9} & 76.9 \\
\hline
\fontsize{8}{0} \selectfont \textbf{NLST-LongCT}  & all & 35.0 & 23.2 & \textbf{36.8} & 27.0 & 40.4 & 32.7 & 43.6 & \textbf{48.7}\\
\hline
\end{tabular}%
\end{table}

\begin{table}[!ht]
\centering
\caption{Segmentation performance (DSC for LongCT and UniToChest; IoU for NLST-Sybil; hit rate for NLST-LongCT) on the test datasets, compared with TotalSegmentator (TS), NoduleNet (NNet) and MedSAM2 (MS2). The segmentation model got prompt annotations (points "P" or boxes "B") from the TM, merged according to (S0), (S1) or (S2) strategies. For the NLST-LongCT, S1 and S2 aggregation strategies were not applicable, as nodule annotations are available for only a single time point.
Bold and underlined values denote the best and second-best performance, respectively. }\label{tab2}
\begin{tabular}{|c|c|cc|cc||cc||cc|}
\cline{3-10}
\multicolumn{2}{c|}{} 
& \multicolumn{4}{c||}{\fontsize{8}{9}\selectfont \textbf{DSC \(\uparrow\)}}
& \multicolumn{2}{c||}{\fontsize{8}{9}\selectfont \textbf{IoU \(\uparrow\)}}
& \multicolumn{2}{c|}{\fontsize{8}{9}\selectfont \textbf{Hit rate \(\uparrow\)}} \\
\cline{3-10}
\multicolumn{2}{c|}{} 
& \multicolumn{2}{c|}{\fontsize{8}{9}\selectfont \textbf{LongCT}}
& \multicolumn{2}{c||}{\fontsize{8}{9}\selectfont \textbf{UniToChest}}
& \multicolumn{2}{c||}{\fontsize{8}{9}\selectfont \textbf{NLST-Sybil}}
& \multicolumn{2}{c|}{\fontsize{8}{9}\selectfont \textbf{NLST-LongCT}} \\
\hline
\multicolumn{2}{|c|}{\fontsize{8}{9}\selectfont \textbf{Model}} 
& \fontsize{8}{9}\selectfont $<1\;\text{cm}$
& \fontsize{8}{9}\selectfont $\geq 1\;\text{cm}$
& \fontsize{8}{9}\selectfont $<1\;\text{cm}$
& \fontsize{8}{9}\selectfont $\geq 1\;\text{cm}$
& \fontsize{8}{9}\selectfont $<1\;\text{cm}$
& \fontsize{8}{9}\selectfont $\geq 1\;\text{cm}$
& \fontsize{8}{9}\selectfont $<1\;\text{cm}$
& \fontsize{8}{9}\selectfont $\geq 1\;\text{cm}$ \\
\hline

\multicolumn{2}{|c|}{TS~\cite{wasserthal2023}} & \fontsize{8}{9}\selectfont 58.8$\pm$0.8 & \fontsize{8}{9}\selectfont\textbf{80.2}$\pm$0.9 & \fontsize{8}{9}\selectfont 54.4$\pm$1.8 & \fontsize{8}{9}\selectfont 74.1$\pm$1.6 & \fontsize{8}{9}\selectfont 13.7$\pm$3.2 & \fontsize{8}{9}\selectfont 17.0$\pm$1.3 & \fontsize{8}{9}\selectfont 78.0$\pm$8.0 & \fontsize{8}{9}\selectfont 12.1$\pm$3.8 \\
\hline

\multicolumn{2}{|c|}{NNet\cite{tang2019}} & \fontsize{8}{9}\selectfont 53.5$\pm$0.9 & \fontsize{8}{9}\selectfont 56.1$\pm$2.0 & \fontsize{8}{9}\selectfont 50.8$\pm$2.1 & \fontsize{8}{9}\selectfont 59.4$\pm$2.6 & \fontsize{8}{9}\selectfont 45.3$\pm$4.5& \fontsize{8}{9}\selectfont 33.6$\pm$1.5&  \fontsize{8}{9}\selectfont 50.2$\pm6.1$& \fontsize{8}{9}\selectfont \textbf{32.8}$\pm8.2$\\
\hline

\multicolumn{2}{|c|}{S0-B} & \fontsize{8}{9}\selectfont 73.4$\pm$1.1 & \fontsize{8}{9}\selectfont 60.4$\pm$2.1 & \fontsize{8}{9}\selectfont 66.1$\pm$1.6 & \fontsize{8}{9}\selectfont 66.0$\pm$2.1 & \fontsize{8}{9}\selectfont\underline{45.9}$\pm$5.0 & \fontsize{8}{9}\selectfont 36.0$\pm$1.7 & \fontsize{8}{9}\selectfont\underline{91.0}$\pm$4.6 & \fontsize{8}{9}\selectfont\underline{24.6}$\pm$5.5 \\

\multicolumn{2}{|c|}{S0-P} & \fontsize{8}{9}\selectfont\textbf{77.6}$\pm$0.8 & \fontsize{8}{9}\selectfont 79.3$\pm$1.6 & \fontsize{8}{9}\selectfont\textbf{67.1}$\pm$1.8 & \fontsize{8}{9}\selectfont\textbf{77.4}$\pm$1.7 & \fontsize{8}{9}\selectfont 44.3$\pm$4.0 & \fontsize{8}{9}\selectfont 33.6$\pm$1.7 & \fontsize{8}{9}\selectfont\textbf{91.0}$\pm$4.6 & \fontsize{8}{9}\selectfont22.5$\pm$5.3 \\

\hline
\multicolumn{2}{|c|}{S1-B} & \fontsize{8}{9}\selectfont 69.4$\pm$1.3 & \fontsize{8}{9}\selectfont 59.2$\pm$2.1 & \fontsize{8}{9}\selectfont 63.5$\pm$2.1 & \fontsize{8}{9}\selectfont 63.3$\pm$2.2 & \fontsize{8}{9}\selectfont\textbf{46.7}$\pm$5.0 & \fontsize{8}{9}\selectfont\underline{36.5}$\pm$1.8 & - & - \\

\multicolumn{2}{|c|}{S1-P} & \fontsize{8}{9}\selectfont\underline{75.2}$\pm$0.9 & \fontsize{8}{9}\selectfont 79.9$\pm$1.3 & \fontsize{8}{9}\selectfont\underline{66.5}$\pm$1.8 & \fontsize{8}{9}\selectfont\underline{77.3}$\pm$1.7 & \fontsize{8}{9}\selectfont 45.0$\pm$4.0 & \fontsize{8}{9}\selectfont 35.2$\pm$1.7 & - & - \\

\hline
\multicolumn{2}{|c|}{S2-B} & \fontsize{8}{9}\selectfont 67.9$\pm$1.3 & \fontsize{8}{9}\selectfont 57.4$\pm$2.2 & \fontsize{8}{9}\selectfont 63.2$\pm$2.1 & \fontsize{8}{9}\selectfont 60.8$\pm$2.2 & \fontsize{8}{9}\selectfont 35.9$\pm$6.6 & \fontsize{8}{9}\selectfont\textbf{37.2}$\pm$1.7 & - & - \\

\multicolumn{2}{|c|}{S2-P} & \fontsize{8}{9}\selectfont 74.1$\pm$0.9 & \fontsize{8}{9}\selectfont\underline{80.1}$\pm$1.3 & \fontsize{8}{9}\selectfont 66.0$\pm$1.7 & \fontsize{8}{9}\selectfont 76.4$\pm$1.8 & \fontsize{8}{9}\selectfont 45.3$\pm$4.6 & \fontsize{8}{9}\selectfont 35.0$\pm$1.6 & - & - \\
\hline
\hline
\multicolumn{2}{|c|}{MS2~\cite{ma2025medsam2}} & \fontsize{8}{9}\selectfont 87.7$\pm$0.4 & \fontsize{8}{9}\selectfont 89.0$\pm$0.6 & \fontsize{8}{9}\selectfont 79.1$\pm$1.2 & \fontsize{8}{9}\selectfont 82.2$\pm$1.3 & \fontsize{8}{9}\selectfont 86.6$\pm$3.1 & \fontsize{8}{9}\selectfont 90.6$\pm$1.0 & \fontsize{8}{9}\selectfont 97.9$\pm$2.1 & \fontsize{8}{9}\selectfont 80.2$\pm$5.7 \\
\hline
\end{tabular}
\end{table}

\textbf{SM performance:} Table~\ref{tab2} reports the segmentation performance of the proposed framework across the four test datasets. Results are reported separately for two nodule diameter ranges ($D \leq 1\;\text{cm}$ and $D > 1\;\text{cm}$), to account for differences between small and large nodules. 
On Longitudinal-CT and UniToChest, the point-based prompting strategy (P) outperformed the box-based counterpart (B) across all temporal aggregation strategies, with the gap being most pronounced for large nodules. 
For instance, on Longitudinal-CT, S0-P achieved a DSC of 79.3\% for large nodules compared to 60.4\% for S0-B; on UniToChest, the gap was even wider (77.4\% vs.\ 66.0\%).
This likely reflects the sensitivity of box-guided segmentation to bounding box precision, as imprecise boxes may include surrounding tissue or miss lesion margins, reducing segmentation accuracy. 
This advantage did not generalize to NLST subsets, where box-based strategies achieved comparable or marginally higher performance (e.g., S0-B: 45.9\% vs.\ S0-P: 44.3\% IoU on NLST-Sybil for small nodules; S0-B: 24.6\% vs.\ S0-P: 22.5\% hit rate on NLST-LongCT for large nodules).
This discrepancy arises from metric sensitivity: DSC penalizes spatial misalignments between predicted and reference masks, whereas IoU and hit rate are less affected by prompt shifts.

Regarding the longitudinal aggregation strategies, the temporal propagation yielded a benefit for large nodules across datasets. 
On Longitudinal-CT, S2-P achieved the highest DSC for large nodules (80.1\%), improving over S0-P (79.3\%) and S1-P (79.9\%), suggesting that aggregating detections from all available time points can refine localization for large, temporally stable nodules. 
A similar trend was observed on UniToChest, where S1-P and S2-P achieved comparable performance to S0-P for large nodules (77.3\% and 76.4\% vs.\ 77.4\%), while on NLST-Sybil the S2-B strategy achieved the highest IoU for large nodules (37.2\%).
For small nodules, longitudinal propagation provided limited benefit.
For example, on Longitudinal-CT, S0-P achieved the highest DSC for small nodules (77.6\%), with S2-P yielding a slightly lower score (74.1\%). 
Small lesions exhibit temporal variability that weakens cross-scan correspondence and limits the effectiveness of longitudinal localization refinement.
  
Comparison with TotalSegmentator reveals a dataset-dependent pattern.
On Longitudinal-CT, TS achieved the highest DSC for large nodules (80.2\%), while our framework outperformed it for small nodules (e.g., S0-P: 77.6\% vs.\ TS: 58.8\%), suggesting that TS generalizes less effectively to sub-centimeter lesions.
On UniToChest, our framework outperformed TS across both size categories (e.g., S0-P: 77.4\% vs.\ TS: 74.1\% for large nodules).
The gap was most pronounced on the NLST subsets, where our framework outperformed TS in IoU on NLST-Sybil (e.g., S0-B: 45.9\% vs.\ TS: 13.7\% for small nodules; 36.0\% vs.\ 17.0\% for large nodules) and hit rate on NLST-LongCT (e.g., S0-B: 91.0\% vs.\ TS: 78.0\% for small nodules).
Compared to NoduleNet, our framework achieved substantially higher DSC on Longitudinal-CT and UniToChest across both size categories (e.g., S0-P: 77.6\% vs.\ 53.5\% for small nodules on Longitudinal-CT; 77.4\% vs.\ 59.4\% for large nodules on UniToChest). 
On NLST-Sybil, performance was comparable, with NNet achieving 45.3\% IoU for small nodules against 46.7\% for S1-B. 
The main exception was hit rate on NLST-LongCT for large nodules, where NNet outperformed our framework (32.8\% vs.\ 24.6\%); for small nodules, however, our framework recovered the lead (91.0\% vs.\ 50.2\%).

MedSAM2, evaluated under an oracle setting with ground-truth annotations as prompts, is included as an approximate upper bound on prompt-driven segmentation performance. It achieved the highest DSC on Longitudinal-CT (87.7\% and 89.0\%) and UniToChest (79.1\% and 82.2\%); the highest hit rate on NLST-LongCT (97.9\% and 80.2\%); and the highest IoU on NLST-Sybil (86.6\% and 90.6\%).

\section{Conclusion}

NoduleFlow provides automatic lung nodule detection, classification, and segmentation from CT series, without requiring prior annotations at inference time. 
The ensemble detector achieved the highest sensitivity across datasets, confirming the benefit of integrating complementary models. 
The segmentation module demonstrated stable performance across test datasets. 
Point-based outperformed box-based prompting, highlighting the sensitivity of box-guided segmentation to bounding box precision. 
Longitudinal aggregation enhanced results for large nodules, while offering limited benefit for small, temporally variable lesions.

Detection performance was strongly dependent on the number and size of lesions. Detection of small metastatic lesions remains a key limitation. 
Future improvements could be achieved by fine-tuning dedicated detection models tailored to metastasis and small nodule distributions.
The study is constrained by the limited availability of longitudinal, publicly annotated datasets. 
Despite these challenges, evaluation across four independent cohorts demonstrates the robustness of the proposed framework to inter-dataset variability. This is particularly noteworthy as most existing methods are validated within a single dataset, limiting their generalizability.
Code will be published upon acceptance.

\end{document}